\documentclass[lettersize,journal]{IEEEtran}
\usepackage{amsmath,amsfonts}
\usepackage{algorithmic}
\usepackage{algorithm}
\usepackage{array}
\usepackage[caption=false,font=normalsize,labelfont=sf,textfont=sf]{subfig}
\usepackage{textcomp}
\usepackage{stfloats}
\usepackage{url}
\usepackage{verbatim}
\usepackage{graphicx}
\usepackage{cite}
\usepackage{multirow}
\begin{document}

\title{Quality-Aware Prototype Memory for Face Representation Learning}

\author{Evgeny Smirnov, Vasiliy Galyuk and Evgeny Lukyanets
\thanks{E. Smirnov, V. Galyuk are with Speech Technology Center, E. Lukyanets is with ITMO University. Corresponding author: Evgeny Smirnov, E-mail: evgeny.versus.smirnov@gmail.com}}

\markboth{Preprint}%
{Shell \MakeLowercase{\textit{et al.}}: Quality-Aware Prototype Memory for Face Representation Learning}


\maketitle

\begin{abstract}
Prototype Memory is a powerful model for face representation learning. It enables training face recognition models on datasets of any size by generating prototypes (classifier weights) on the fly and efficiently utilizing them. Prototype Memory demonstrated strong results in many face recognition benchmarks. However, the algorithm of prototype generation, used in it, is prone to the problems of imperfectly calculated prototypes in case of low-quality or poorly recognizable faces in the images, selected for the prototype creation. All images of the same person presented in the mini-batch are used with equal weights, and the resulting averaged prototype can be contaminated by imperfect embeddings of low-quality face images. This may lead to misleading training signals and degrade the performance of the trained models. In this paper, we propose a simple and effective way to improve Prototype Memory with quality-aware prototype generation. Quality-Aware Prototype Memory uses different weights for images of different quality in the process of prototype generation. With this improvement, prototypes receive more informative signals from high-quality images and are less affected by low-quality ones. We propose and compare several methods of quality estimation and usage, perform extensive experiments on the different face recognition benchmarks and demonstrate the advantages of the proposed model compared to the basic version of Prototype Memory.
\end{abstract}


\section{Introduction}
\IEEEPARstart{F}{ace} recognition is an important technology in the modern world \cite{dfrsurvey, du2020elements, wang2022survey}. The main components of its current success are deep neural architectures \cite{hinton2015deep}, large-scale training datasets \cite{an2020partial, zhu2021webface260m} and various specialized face representation learning methods \cite{deng2019arcface, elasticface, pm, an2022killing}. Face recognition research directions include the areas, where current models still exhibit limitations: large pose and age variations \cite{cfpfp, cplfw, agedb, calfw}, ethnicity and demographic bias \cite{RFW}, unconstrained and difficult face image conditions \cite{ijbc, singh2019recognizing}.

One of the ways to overcome these problems is usage of large face image datasets for training \cite{schroff2015facenet, zhu2021webface260m}, containing millions of persons and tens or hundreds of millions of images. Softmax-based methods, such as ArcFace \cite{deng2019arcface}, AdaFace \cite{kim2022adaface} and ElasticFace \cite{elasticface}, are among the most effective approaches for training face recognition models, must be adapted for datasets of such size. For this purpose, there were proposed several sampling-based methods of training: Positive Plus Randomly Negative (PPRN) \cite{an2020partial}, D-Softmax-K \cite{he2020softmax} and Prototype Memory \cite{pm}. With Prototype Memory, for example, the training could be performed on face datasets with unlimited number of persons and images.

\begin{figure} [t]
	\begin{center}
		\includegraphics[width=1.0\linewidth]{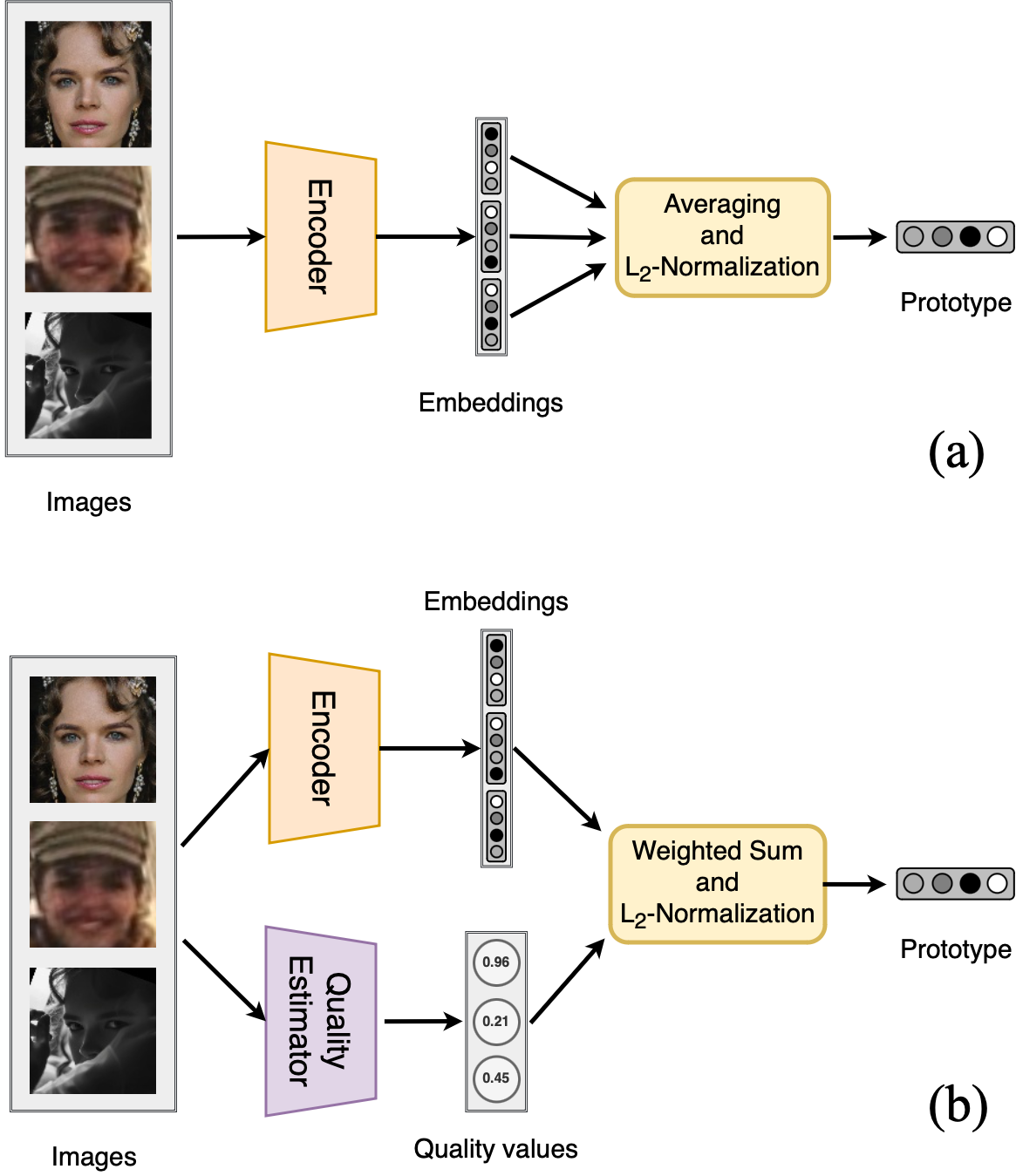}
	\end{center}
	\caption{Prototype generation with (a) Prototype Memory, (b) Quality-Aware Prototype Memory}
	\label{fig:qpm-front}
\end{figure}

Another direction of research in face recognition area is face quality estimation \cite{schlett2022face, merkle2022state}. Information about the quality of face images could be used to filter unrecognizable faces and improve the performance of face recognition models. Face quality and recognizability labels also could be used for learning better face representations \cite{saadabadi2023quality, kim2022adaface} with more precise training signals and less noisy gradients.

In this paper we propose the way to include face quality values into the process of Prototype Memory-based training of face representations. Proposed Quality-Aware Prototype Memory performs better, than basic version, on different face recognition benchmarks. We also compare several methods of face quality estimation, which could be used.

\section{Related work}

\subsection{Face recognition}
The typical face recognition pipeline consists of detecting faces in images using a face detector \cite{deng2020retinaface, qi2022yolo5face}, performing alignment, cropping and preprocessing \cite{timoshenko2014effective, deng2023harnessing}, sometimes - liveness detection \cite{melnikov2015audiovisual, luckyanets2017bimodal, lavrentyeva2018interactive, timoshenko2019large} and then computing a face representation using an encoder network. The result is in a form of $L_2$-normalized vector. These vectors (embeddings) can be used for face verification and identification by calculating cosine similarities between images. Embeddings of different images of the same person should be close to each other, embeddings of the images of different persons should be far. Training of encoder neural network is performed with specialized methods like ArcFace \cite{deng2019arcface}, ElasticFace \cite{elasticface} and Prototype Memory \cite{pm}, which provide appropriate training signals. To get better results, methods of hard example mining \cite{dm, ae, cmb} and data augmentation \cite{facemix, zeno2021pfa} could be used.

\subsection{Face quality estimation}
Not all face images are created equal, they have different image quality, different face poses, occlusions, attributes, lighting conditions, etc. Sometimes, the face in image could be detected, but could not be reliably recognized. To filter such images, and to measure the quality and the recognizability of faces, one could use face quality estimation methods \cite{schlett2022face} such as CR-FIQA \cite{boutros2023cr}, FaceQAN \cite{babnik2022faceqan}, L2RT-FIQA \cite{chen2022l2rt}, DifFIQA \cite{babnik2023diffiqa} and others \cite{najafzadeh2023face, babnik2022assessing, huber2022evaluating, ou2023troubleshooting, okcu2020efficient, deng2023harnessing, babnik2023optimization, grimmer2023pose, huoa2023unsupervised}. Sometimes these methods are inserted in the process of face representation learning \cite{medvedev2022towards, saadabadi2023quality, chai2023recognizability, meng2021magface, terhorst2023qmagface, kim2022adaface, lv2022hq2cl, wang2022cqa, EPF}, improving the results with more precise training signals. However, none of these methods are used in the Prototype Memory-based face representation learning.

\section{Proposed method}
We propose to use face quality values in the process of Prototype Memory's generation of prototypes, improving their ability to represent corresponding identities, and therefore the training signals they provide. The general scheme of the proposed prototype generation method is shown in Fig. 1.

\subsection{Motivation}
Prototype Memory \cite{pm} is a model for learning face representations. It consists of a limited-size memory module for storing class prototypes and employs a set of algorithms to update it efficiently. New class prototypes are generated on the fly using exemplar embeddings in the current mini-batch. These prototypes are enqueued to the memory and used in a role of classifier weights for softmax classification-based training. Prototypes are regularly refreshed, and oldest ones are dequeued and disposed of. Prototype Memory proved its effectiveness, but it has some weaknesses. One of them is the way new prototypes are generated:

\begin{equation}
\mathbf{p}_{new} = F_{norm}(\frac{\sum_{j=1}^{k} \mathbf{x}_{j}}{k}),
\end{equation}

$\mathbf{x}_{j}$ is the $j$-th exemplar embedding, produced by encoder, $k$ is the number of images, used for prototype generation, and $F_{norm}$ is $L_2$-normalization function. The resulting vector $\mathbf{p}_{new}$ is the new class prototype.

All images, used for prototype generation, contribute equally to the final prototype position. If the images (and their embeddings) are of high quality, then the resulting prototype will accurately represent the corresponding person in the embedding space, and the training signals, produced using these prototypes, will be precise enough to train good face recognition network. However, if some of the images are of low quality, or the faces in some of these images are not recognizable, then their influence will position resulting prototypes in a wrong place in the embedding space (see Fig. 2), having a misleading effect on the direction of the training signals.

\begin{figure} [t]
	\begin{center}
		\includegraphics[width=1.0\linewidth]{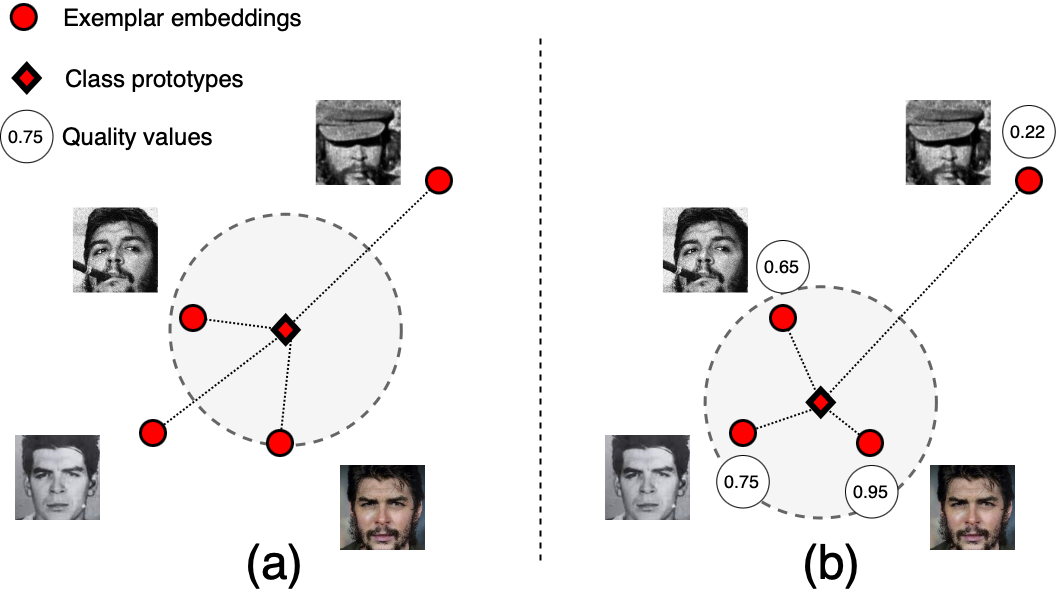}
	\end{center}
	\caption{Example prototypes, generated with (a) Prototype Memory, (b) Quality-Aware Prototype Memory.}
	\label{fig:qpm-example}
\end{figure}

\subsection{Quality-Aware Prototype Memory}
To prevent the destructive effects from misplaced prototypes, we propose to use face quality values as weights in prototype generation process:

\begin{equation}
\mathbf{p}_{new} = F_{norm}\left(\frac{\sum_{j=1}^{k} \mathit{q}_{j} \mathbf{x}_{j}}{\sum_{j=1}^{k} \mathit{q}_{j}}\right),
\end{equation}

$\mathit{q}_{j}$ is the $j$-th exemplar quality value, produced by some quality estimator, $\mathbf{x}_{j}$ is the $j$-th exemplar embedding, produced by encoder, $k$ is the number of images, used for prototype generation, and $F_{norm}$ is $L_2$-normalization function. The resulting vector $\mathbf{p}_{new}$ is the new quality-aware class prototype.

In this way, the influence of low-quality or unrecognizable faces will be reduced, leading to prototypes that produce more accurate training signals.

Quality values could be estimated using many different methods, both with separate models and directly using features of the current face recognition network. In the next section we propose a couple of simple ways to get face quality values using the latter option with very moderate extra computation efforts.

\subsection{Face quality estimation methods}
\textbf{Feature norms}, taken from face embeddings before performing $L_2$-normalization, could be used as a way to approximate the quality of face images \cite{meng2021magface, kim2022adaface}. Higher quality faces tend to possess larger feature norm values. For Quality-Aware Prototype Memory we simply use the feature norm values, divided by the maximum feature norm value in the mini-batch:

\begin{equation}
\mathit{q}_{j} = \frac{\|\mathbf{x}_{j}\|}{\|\mathbf{x}\|_{max}},
\end{equation}

\textbf{Recognizability scores}, calculated using cosine distances from "unrecognizable identity" \cite{deng2023harnessing, stc_odyssey}, is another way to measure face quality. For this case, we create ${\mathbf{p}}_{UI}$ ("unrecognizable identity prototype") in a Prototype Memory, and "refresh" it every ${N}_{upd}$ iterations using standard Prototype Memory algorithm for and a set of images with "unrecognizable faces". To estimate face quality values for a mini-batch examples, we compute cosine distances between embeddings and ${\mathbf{p}}_{UI}$, and use the formula from \cite{deng2023harnessing}:

\begin{equation}
\mathit{q}_{j} = ({D}_{cos}({\mathbf{x}}_{j},{{\mathbf{p}}_{UI}}))^2
\end{equation}

where ${D}_{cos}$ is cosine distance. We also performed experiments with "hard" version of recognizability scores, where only one embedding with the largest recognizability score is used as a new prototype.

\section{Experiments}
To confirm the effectiveness of our method, we conducted experiments with proposed methods on several face recognition benchmarks.

\subsection{Implementation details}
\textbf{Datasets}. For training we used Glint360k-R \cite{pm} dataset with $353,658$ persons and $16,4$ million images, which is a cleaned version of Glint360k \cite{an2020partial} dataset.
For testing, we used CFP-FP \cite{cfpfp}, AgeDB-30 \cite{agedb}, CALFW \cite{calfw}, CPLFW \cite{cplfw}, XQLFW \cite{xqlfw}, RFW \cite{RFW}, MegaFace \cite{kemelmacher2016megaface} (refined version \cite{deng2019arcface}) and IJB-C \cite{ijbc} datasets.

\textbf{Training Settings}. For the experiments we used ResNet-34 \cite{he2016deep} as an encoder architecture. We trained the nets with Prototype Memory \cite{pm} and Quality-Aware Prototype Memory with embedding size $d = 256$, memory size $m = 100,000$ and refresh ratio $r = 0.2$. Composite mini-batch \cite{cmb} of size $512$ was used, with $384$ images sampled by group-based iterate-and-shuffle \cite{pm} strategy, and $128$ images - using a combination of Multi-Doppelganger Mining \cite{pm} and Hardness-aware example mining \cite{pm}, with $h = 0.25$, $2$ classes sampled at random and $30$ classes sampled using doppelgangers. For all experiments we used FaceMix-BI \cite{facemix} data augmentation with $5\%$ probability of both Intra-Class and Background FaceMix. We compared the performance of proposed method with different loss functions - CosFace \cite{wang2018cosface} with $m = 0.4$ and $s = 64$, ArcFace \cite{deng2019arcface} with $m = 0.5$ and $s = 64$, ElasticFace-Cos+ \cite{elasticface} with $m = 0.4$, $\sigma = 0.025$ and $s = 64$, and ElasticFace-Arc+ \cite{elasticface} with $m = 0.5$, $\sigma = 0.0175$ and $s = 64$. In the experiments with recognizability-based quality scores, we refreshed ${\mathbf{p}}_{UI}$ every ${N}_{upd} = 100$ iterations. The initial learning rate was set to $0.1$ and reduced by a factor of $10$ at $100K$, $200K$, $250K$, and $275K$ iterations, in total training was performed for $300K$ iterations.

To perform large-scale experiment, we used ResNet-100 \cite{he2016deep} network, pre-trained on Glint360k dataset. We fine-tuned it with different Prototype Memory variants for $200k$ more iterations on Glint360k-R dataset using batch size of $512$, learning rate starting with $0.01$ and reduced by a factor of $10$ at $50k$, $125k$ and $175k$ iterations. We used CosFace with $m = 0.4$ and $s = 64$ for Prototype Memory and Quality-Aware Prototype Memory (with recognizability-based quality values) with $d = 512$, $M = 200,000$ and $r = 0.2$, and composite mini-batch with the same parameters, as in previous experiments.

To obtain the final accuracy values, we averaged the results obtained from the last five snapshots ($260k$, $270k$, $280k$, $290k$ and $300k$ iterations for ResNet-34 nets, and $180k$, $185k$, $190k$, $195k$, $200k$ - for ResNet-100 nets).

\subsection{Results}
Results of experiments with different variants of face quality estimation for QA-Prototype Memory (R-34) are presented in Table I and Table II. (N) means "feature norm"-based estimation, (H) - "hard" recognizability-based estimation, and (S) - "soft" (usual) recognizability-based estimation.

\begin{table}[h]
	\caption{Experiments with different variants of face quality estimation for QA-Prototype Memory (R-34). Results on CFP-FP, AgeDB-30, CALFW, CPLFW and XQLFW datasets}
	\begin{center}
		\label{table:r34diff}
		\begin{tabular}{|c|ccccc|}
			\hline
			Method & CFP-FP & AgeDB & CALFW & CPLFW & XQLFW \\
			\hline\hline
			PM & 98.47 & 97.90 & 95.62 & 92.95 & 89.72 \\
                QA-PM (N) & 98.39 & 97.85 & \textbf{95.68} & 93.01 & \textbf{90.35}\\
			QA-PM (H) & 98.47 & 97.73 & 95.61 & 92.74 & 89.37\\
                QA-PM (S) & \textbf{98.62} & \textbf{97.98} & 95.59 & \textbf{93.43} & 90.14 \\
			\hline
		\end{tabular}
	\end{center}
\end{table}
 
\begin{table}[h]
	\caption{Experiments with different variants of face quality estimation for QA-Prototype Memory (R-34). Results on RFW dataset}
	\begin{center}
		\label{table:r34rfw}
		\begin{tabular}{|c|cccc|c|}
			\hline
			Method & African & Asian & Indian & Caucasian & Average \\
			\hline\hline
			PM & 92.52 & 93.25 & 95.09 & 96.90 & 94.13 \\
                QA-PM (N) & 92.22 & 93.04 & 95.15 & 97.12 & 94.13 \\
			QA-PM (H) & 91.81 & 92.86 & 94.72 & 96.79 & 93.85 \\
                QA-PM (S) & \textbf{92.72} & \textbf{93.87} & \textbf{95.44} & \textbf{97.31} & \textbf{94.56} \\
			\hline
		\end{tabular}
	\end{center}
\end{table}

Results of experiments with different loss functions, applied to QA-Prototype Memory, are presented in Table III.

\begin{table}[h]
	\caption{Experiments with QA-Prototype Memory (R-34) and different loss functions. Results on RFW and MegaFace (R) datasets}
	\begin{center}
		\label{table:r34loss}
		\begin{tabular}{|c|c|cc|}
			\hline
			\multirow{2}{*}{Method} & \multirow{2}{*}{RFW (Average)} & \multicolumn{2}{|c|}{MegaFace (R)} \\
                \cline{3-4}
			&& Id & Ver\\
			\hline\hline
			PM (CosFace) & 94.13 & 97.28 & 97.00 \\
   			QA-PM (CosFace) & \textbf{94.56} & \textbf{97.43} & \textbf{97.41} \\
                \hline
			PM (ArcFace) & 94.07 & 97.29 & 97.29 \\
   			QA-PM (ArcFace) & \textbf{94.35} & \textbf{97.31} & \textbf{97.36} \\
                \hline
			PM (ElasticFace-Cos+) & 94.18 & 97.28 & 97.07 \\
   			QA-PM (ElasticFace-Cos+) & \textbf{94.21} & \textbf{97.45} & \textbf{97.39} \\
                \hline
			PM (ElasticFace-Arc+) & \textbf{94.31} & 97.35 & 97.34 \\
   			QA-PM (ElasticFace-Arc+) & 94.08 & \textbf{97.41} & \textbf{97.55} \\
			\hline
		\end{tabular}
	\end{center}
\end{table}

Results of the experiments with large networks (R-100) are presented in Table IV and Table V.

\begin{table}[h]
	\caption{Experiments with QA-Prototype Memory (R-100) on RFW dataset}
	\begin{center}
		\label{table:r100rfw}
		\begin{tabular}{|c|cccc|c|}
			\hline
			Method & African & Asian & Indian & Caucasian & Average \\
			\hline\hline
			PM & 97.72 & 97.99 & 98.34 & \textbf{99.05} & 98.23 \\
			QA-PM & \textbf{97.81} & \textbf{98.06} & \textbf{98.40} & 98.96 & \textbf{98.29}\\
			\hline
		\end{tabular}
	\end{center}
\end{table}

\begin{table}[h]
	\caption{Experiments with QA-Prototype Memory (R-100) on XQLFW, IJB-C and MegaFace (R) datasets}
	\begin{center}
		\label{table:r100xim}
		\begin{tabular}{|c|c|ccc|cc|}
			\hline
			\multirow{2}{*}{Method} & \multirow{2}{*}{XQLFW} & \multicolumn{3}{|c|}{IJB-C} & \multicolumn{2}{|c|}{MegaFace (R)} \\
                \cline{3-7}
			&& 1e-2 & 1e-3 & 1e-4 & Id & Ver\\
			\hline\hline
			PM & 91.76 & 98.62 & 98.02 & 97.17 & 98.45 & 98.46 \\
			QA-PM & \textbf{92.25} & \textbf{98.86} & \textbf{98.26} & \textbf{97.20} & \textbf{98.55} & \textbf{98.61} \\
			\hline
		\end{tabular}
	\end{center}
\end{table}

As we can see from the experiments, Quality-Aware Prototype Memory consistently performs better, than basic variant of Prototype Memory. The best way of face quality estimation appears to be using "soft" recognizability scores.

\section{Conclusion}
In this paper we proposed Quality-Aware Prototype Memory - a simple improvement over Prototype Memory. This face representation learning model uses face quality scores in the process of prototype generation. With the help of these scores, prototypes are generated more accurately, even in the presence of low-quality or unrecognizable face images. Better prototypes lead to more accurate training signals, which are essential for efficient representation learning. We proposed a number of ways to compute face quality scores, and performed experiments with different variants of Quality-Aware Prototype memory using various face recognition benchmarks. Achieved results demonstrate the superiority of the proposed method.


{\small
\bibliographystyle{IEEEtran}
\bibliography{vs}

\begin{thebibliography}{10}
\providecommand{\url}[1]{#1}
\csname url@samestyle\endcsname
\providecommand{\newblock}{\relax}
\providecommand{\bibinfo}[2]{#2}
\providecommand{\BIBentrySTDinterwordspacing}{\spaceskip=0pt\relax}
\providecommand{\BIBentryALTinterwordstretchfactor}{4}
\providecommand{\BIBentryALTinterwordspacing}{\spaceskip=\fontdimen2\font plus
\BIBentryALTinterwordstretchfactor\fontdimen3\font minus
  \fontdimen4\font\relax}
\providecommand{\BIBforeignlanguage}[2]{{%
\expandafter\ifx\csname l@#1\endcsname\relax
\typeout{** WARNING: IEEEtran.bst: No hyphenation pattern has been}%
\typeout{** loaded for the language `#1'. Using the pattern for}%
\typeout{** the default language instead.}%
\else
\language=\csname l@#1\endcsname
\fi
#2}}
\providecommand{\BIBdecl}{\relax}
\BIBdecl

\bibitem{dfrsurvey}
I.~Masi, Y.~Wu, T.~Hassner, and P.~Natarajan, ``Deep face recognition: A
  survey,'' \emph{Neurocomputing}, 2021.

\bibitem{du2020elements}
H.~Du, H.~Shi, D.~Zeng, and T.~Mei, ``The elements of end-to-end deep face
  recognition: A survey of recent advances,'' \emph{arXiv preprint
  arXiv:2009.13290}, 2020.

\bibitem{wang2022survey}
X.~Wang, J.~Peng, S.~Zhang, B.~Chen, Y.~Wang, and Y.~Guo, ``A survey of face
  recognition,'' \emph{arXiv preprint arXiv:2212.13038}, 2022.

\bibitem{hinton2015deep}
G.~Hinton, Y.~LeCun, and Y.~Bengio, ``Deep learning,'' \emph{Nature}, vol. 521,
  no. 7553, pp. 436--444, 2015.

\bibitem{an2020partial}
X.~An, X.~Zhu, Y.~Xiao, L.~Wu, M.~Zhang, Y.~Gao, B.~Qin, D.~Zhang, and Y.~Fu,
  ``{Partial FC: Training 10 Million Identities on a Single Machine},''
  \emph{arXiv preprint arXiv:2010.05222}, 2020.

\bibitem{zhu2021webface260m}
Z.~Zhu, G.~Huang, J.~Deng, Y.~Ye, J.~Huang, X.~Chen, J.~Zhu, T.~Yang, J.~Lu,
  D.~Du, and J.~Zhou, ``{WebFace260M: A Benchmark Unveiling the Power of
  Million-Scale Deep Face Recognition},'' in \emph{CVPR}, 2021.

\bibitem{deng2019arcface}
J.~Deng, J.~Guo, N.~Xue, and S.~Zafeiriou, ``Arcface: Additive angular margin
  loss for deep face recognition,'' in \emph{Proceedings of the IEEE/CVF
  conference on CVPR}, 2019, pp. 4690--4699.

\bibitem{elasticface}
F.~Boutros, N.~Damer, F.~Kirchbuchner, and A.~Kuijper, ``{ElasticFace}: Elastic
  margin loss for deep face recognition,'' in \emph{CVPR Workshops}, 2022.

\bibitem{pm}
E.~Smirnov, N.~Garaev, V.~Galyuk, and E.~Lukyanets, ``Prototype memory for
  large-scale face representation learning,'' \emph{IEEE Access}, vol.~10, pp.
  12\,031--12\,046, 2022.

\bibitem{an2022killing}
X.~An, J.~Deng, J.~Guo, Z.~Feng, X.~Zhu, J.~Yang, and T.~Liu, ``Killing two
  birds with one stone: Efficient and robust training of face recognition cnns
  by partial fc,'' in \emph{CVPR}, 2022.

\bibitem{cfpfp}
S.~Sengupta, J.-C. Chen, C.~Castillo, V.~M. Patel, R.~Chellappa, and D.~W.
  Jacobs, ``{Frontal to Profile Face Verification in the Wild},'' in
  \emph{WACV}, 2016.

\bibitem{cplfw}
T.~Zheng and W.~Deng, ``{Cross-Pose LFW: A Database for Studying Cross-Pose
  Face Recognition in Unconstrained Environments},'' \emph{Beijing University
  of Posts and Telecommunications, Tech. Rep}, 2018.

\bibitem{agedb}
S.~Moschoglou, A.~Papaioannou, C.~Sagonas, J.~Deng, I.~Kotsia, and
  S.~Zafeiriou, ``{AgeDB: the first manually collected, in-the-wild age
  database},'' in \emph{CVPR Workshops}, 2017.

\bibitem{calfw}
T.~Zheng, W.~Deng, and J.~Hu, ``Cross-age lfw: A database for studying
  cross-age face recognition in unconstrained environments,'' \emph{arXiv
  preprint arXiv:1708.08197}, 2017.

\bibitem{RFW}
M.~Wang, W.~Deng, J.~Hu, X.~Tao, and Y.~Huang, ``{Racial Faces in the Wild:
  Reducing Racial Bias by Information Maximization Adaptation Network},'' in
  \emph{ICCV}, 2019.

\bibitem{ijbc}
B.~Maze, J.~Adams, J.~A. Duncan, N.~Kalka, T.~Miller, C.~Otto, A.~K. Jain,
  W.~T. Niggel, J.~Anderson, J.~Cheney \emph{et~al.}, ``{IARPA Janus
  Benchmark-C: Face Dataset and Protocol},'' in \emph{ICB}, 2018.

\bibitem{singh2019recognizing}
M.~Singh, R.~Singh, M.~Vatsa, N.~K. Ratha, and R.~Chellappa, ``{Recognizing
  Disguised Faces in the Wild},'' \emph{IEEE Transactions on Biometrics,
  Behavior, and Identity Science}, 2019.

\bibitem{schroff2015facenet}
F.~Schroff, D.~Kalenichenko, and J.~Philbin, ``{FaceNet: A Unified Embedding
  for Face Recognition and Clustering},'' in \emph{CVPR}, 2015.

\bibitem{kim2022adaface}
M.~Kim, A.~K. Jain, and X.~Liu, ``{AdaFace}: Quality adaptive margin for face
  recognition,'' in \emph{CVPR}, 2022.

\bibitem{he2020softmax}
L.~He, Z.~Wang, Y.~Li, and S.~Wang, ``{Softmax Dissection: Towards
  Understanding Intra- and Inter-Class Objective for Embedding Learning},'' in
  \emph{AAAI}, 2020.

\bibitem{schlett2022face}
T.~Schlett, C.~Rathgeb, O.~Henniger, J.~Galbally, J.~Fierrez, and C.~Busch,
  ``Face image quality assessment: A literature survey,'' \emph{ACM Computing
  Surveys (CSUR)}, vol.~54, no. 10s, pp. 1--49, 2022.

\bibitem{merkle2022state}
J.~Merkle, C.~Rathgeb, B.~Tams, D.-P. Lou, A.~D{\"o}rsch, and P.~Drozdowski,
  ``State of the art of quality assessment of facial images,'' \emph{arXiv
  preprint arXiv:2211.08030}, 2022.

\bibitem{saadabadi2023quality}
M.~S.~E. Saadabadi, S.~R. Malakshan, A.~Zafari, M.~Mostofa, and N.~M.
  Nasrabadi, ``A quality aware sample-to-sample comparison for face
  recognition,'' in \emph{WACV}, 2023.

\bibitem{deng2020retinaface}
J.~Deng, J.~Guo, E.~Ververas, I.~Kotsia, and S.~Zafeiriou, ``{RetinaFace:
  Single-Shot Multi-Level Face Localisation in the Wild},'' in \emph{CVPR},
  2020.

\bibitem{qi2022yolo5face}
D.~Qi, W.~Tan, Q.~Yao, and J.~Liu, ``Yolo5face: why reinventing a face
  detector,'' in \emph{European Conference on Computer Vision}.\hskip 1em plus
  0.5em minus 0.4em\relax Springer, 2022, pp. 228--244.

\bibitem{timoshenko2014effective}
D.~Timoshenko, V.~Grishkin, and E.~Smirnov, ``Effective false positive
  reduction in multilevel face detection system using convolutional neural
  networks,'' in \emph{ICCTPEA}, 2014.

\bibitem{deng2023harnessing}
S.~Deng, Y.~Xiong, M.~Wang, W.~Xia, and S.~Soatto, ``Harnessing unrecognizable
  faces for improving face recognition,'' in \emph{WACV}, 2023.

\bibitem{melnikov2015audiovisual}
A.~Melnikov, R.~Akhunzyanov, O.~Kudashev, and E.~Luckyanets, ``Audiovisual
  liveness detection,'' in \emph{ICIAP}, 2015.

\bibitem{luckyanets2017bimodal}
E.~Luckyanets, A.~Melnikov, O.~Kudashev, S.~Novoselov, and G.~Lavrentyeva,
  ``Bimodal anti-spoofing system for mobile security,'' in \emph{SPECOM}, 2017.

\bibitem{lavrentyeva2018interactive}
G.~Lavrentyeva, O.~Kudashev, A.~Melnikov, M.~De~Marsico, and Y.~Matveev,
  ``Interactive photo liveness for presentation attacks detection,'' in
  \emph{ICIAR}.\hskip 1em plus 0.5em minus 0.4em\relax Springer, 2018.

\bibitem{timoshenko2019large}
D.~Timoshenko, K.~Simonchik, V.~Shutov, P.~Zhelezneva, and V.~Grishkin, ``Large
  crowdcollected facial anti-spoofing dataset,'' \emph{CSIT}, 2019.

\bibitem{dm}
E.~Smirnov, A.~Melnikov, S.~Novoselov, E.~Luckyanets, and G.~Lavrentyeva,
  ``{Doppelganger Mining for Face Representation Learning},'' in \emph{ICCV
  Workshops}, 2017.

\bibitem{ae}
E.~Smirnov, A.~Melnikov, A.~Oleinik, E.~Ivanova, I.~Kalinovskiy, and
  E.~Luckyanets, ``{Hard Example Mining with Auxiliary Embeddings},'' in
  \emph{CVPR Workshops}, 2018.

\bibitem{cmb}
E.~Smirnov, A.~Oleinik, A.~Lavrentev, E.~Shulga, V.~Galyuk, N.~Garaev,
  M.~Zakuanova, and A.~Melnikov, ``{Face Representation Learning using
  Composite Mini-Batches},'' in \emph{ICCV Workshops}, 2019.

\bibitem{facemix}
N.~Garaev, E.~Smirnov, V.~Galyuk, and E.~Lukyanets, ``{FaceMix}: Transferring
  local regions for data augmentation in face recognition,'' in \emph{ICONIP},
  2022.

\bibitem{zeno2021pfa}
B.~Zeno, I.~Kalinovskiy, and Y.~Matveev, ``Pfa-gan: Pose face augmentation
  based on generative adversarial network,'' \emph{Informatica}, vol.~32,
  no.~2, pp. 425--440, 2021.

\bibitem{boutros2023cr}
F.~Boutros, M.~Fang, M.~Klemt, B.~Fu, and N.~Damer, ``{CR-FIQA}: face image
  quality assessment by learning sample relative classifiability,'' in
  \emph{CVPR}, 2023, pp. 5836--5845.

\bibitem{babnik2022faceqan}
{\v{Z}}.~Babnik, P.~Peer, and V.~{\v{S}}truc, ``{FaceQAN}: Face image quality
  assessment through adversarial noise exploration,'' in \emph{ICPR}, 2022.

\bibitem{chen2022l2rt}
Z.~Chen and H.~Yang, ``{L2RT-FIQA}: Face image quality assessment via
  learning-to-rank transformer,'' in \emph{International Forum on Digital TV
  and Wireless Multimedia Communications}.\hskip 1em plus 0.5em minus
  0.4em\relax Springer, 2022, pp. 270--285.

\bibitem{babnik2023diffiqa}
{\v{Z}}.~Babnik, P.~Peer, and V.~{\v{S}}truc, ``{DifFIQA}: Face image quality
  assessment using denoising diffusion probabilistic models,'' \emph{arXiv
  preprint arXiv:2305.05768}, 2023.

\bibitem{najafzadeh2023face}
N.~Najafzadeh, H.~Kashiani, M.~S.~E. Saadabadi, N.~A. Talemi, S.~R. Malakshan,
  and N.~M. Nasrabadi, ``Face image quality vector assessment for biometrics
  applications,'' in \emph{WACV}, 2023, pp. 511--520.

\bibitem{babnik2022assessing}
Z.~Babnik and V.~{\v{S}}truc, ``Assessing bias in face image quality
  assessment,'' in \emph{EUSIPCO}, 2022.

\bibitem{huber2022evaluating}
M.~Huber, P.~Terh{\"o}st, F.~Kirchbuchner, N.~Damer, and A.~Kuijper, ``On
  evaluating pixel-level face image quality assessment,'' in \emph{EUSIPCO},
  2022.

\bibitem{ou2023troubleshooting}
F.-Z. Ou, B.~Chen, C.~Li, S.~Wang, and S.~Kwong, ``Troubleshooting ethnic
  quality bias with curriculum domain adaptation for face image quality
  assessment,'' in \emph{ICCV}, 2023.

\bibitem{okcu2020efficient}
S.~B. Okcu, B.~O. {\"O}zkalayc{\i}, and C.~{\c{C}}{\i}{\u{g}}la, ``An efficient
  method for face quality assessment on the edge,'' in \emph{ECCV}, 2020.

\bibitem{babnik2023optimization}
{\v{Z}}.~Babnik, N.~Damer, and V.~{\v{S}}truc, ``Optimization-based improvement
  of face image quality assessment techniques,'' in \emph{IWBF}, 2023.

\bibitem{grimmer2023pose}
M.~Grimmer, C.~Rathgeb, and C.~Busch, ``Pose impact estimation on face
  recognition using 3d-aware synthetic data with application to quality
  assessment,'' \emph{arXiv preprint arXiv:2303.00491}, 2023.

\bibitem{huoa2023unsupervised}
L.~HUOa, Y.~Xiong, J.~SUNb, Y.~NIEa, and W.~SUb, ``Unsupervised face image
  quality assessment based on face recognition,'' \emph{ICMAT}, vol.~33, p.~77,
  2023.

\bibitem{medvedev2022towards}
I.~Medvedev, J.~Tremo{\c{c}}o, B.~Mano, L.~E. Santo, and N.~Gon{\c{c}}alves,
  ``Towards understanding the character of quality sampling in deep learning
  face recognition,'' \emph{IET Biometrics}, vol.~11, no.~5, pp. 498--511,
  2022.

\bibitem{chai2023recognizability}
J.~C.~L. Chai, T.-S. Ng, C.-Y. Low, J.~Park, and A.~B.~J. Teoh,
  ``Recognizability embedding enhancement for very low-resolution face
  recognition and quality estimation,'' in \emph{CVPR}, 2023.

\bibitem{meng2021magface}
Q.~Meng, S.~Zhao, Z.~Huang, and F.~Zhou, ``{MagFace: A Universal Representation
  for Face Recognition and Quality Assessment},'' in \emph{CVPR}, 2021.

\bibitem{terhorst2023qmagface}
P.~Terh{\"o}rst, M.~Ihlefeld, M.~Huber, N.~Damer, F.~Kirchbuchner, K.~Raja, and
  A.~Kuijper, ``{QMagFace}: Simple and accurate quality-aware face
  recognition,'' in \emph{WACV}, 2023.

\bibitem{lv2022hq2cl}
X.~Lv, C.~Yu, H.~Jin, and K.~Liu, ``{HQ2CL}: A high-quality class center
  learning system for deep face recognition,'' \emph{IEEE Transactions on Image
  Processing}, vol.~31, pp. 5359--5370, 2022.

\bibitem{wang2022cqa}
Q.~Wang and G.~Guo, ``{CQA-Face}: Contrastive quality-aware attentions for face
  recognition,'' in \emph{AAAI}, 2022.

\bibitem{EPF}
Y.~Liu, g.~song, m.~zhang, j.~liu, y.~zhou, and j.~yan, ``{Towards
  Flops-Constrained Face Recognition},'' in \emph{ICCV Workshops}, 2019.

\bibitem{stc_odyssey}
G.~Lavrentyeva, S.~Novoselov, V.~Volokhov, A.~Avdeeva, A.~Gusev,
  A.~Vinogradova, I.~Korsunov, A.~Kozlov, T.~Pekhovsky, A.~Shulipa
  \emph{et~al.}, ``{STC speaker recognition system for the NIST SRE 2021},'' in
  \emph{Odyssey}, 2022.

\bibitem{xqlfw}
M.~Knoche, S.~Hormann, and G.~Rigoll, ``Cross-quality {LFW}: A database for
  analyzing cross-resolution image face recognition in unconstrained
  environments,'' in \emph{FG}, 2021.

\bibitem{kemelmacher2016megaface}
I.~Kemelmacher-Shlizerman, S.~M. Seitz, D.~Miller, and E.~Brossard, ``{The
  MegaFace Benchmark: 1 Million Faces for Recognition at Scale},'' in
  \emph{CVPR}, 2016.

\bibitem{he2016deep}
K.~He, X.~Zhang, S.~Ren, and J.~Sun, ``{Deep Residual Learning for Image
  Recognition},'' in \emph{CVPR}, 2016.

\bibitem{wang2018cosface}
H.~Wang, Y.~Wang, Z.~Zhou, X.~Ji, Z.~Li, D.~Gong, J.~Zhou, and W.~Liu,
  ``{CosFace: Large Margin Cosine Loss for Deep Face Recognition},'' in
  \emph{CVPR}, 2018.

\end{thebibliography}
}


\vfill

\end{document}